\documentclass[letterpaper, 10 pt, conference]{ieeeconf}  
\usepackage{times}
\usepackage{color, colortbl}

\newcommand{\hdyn}{f_H}
\newcommand{\hstate}{x_H}

\newcommand{\hctrl}{u_H}

\newcommand{\dt}{\Delta t}
\newcommand{\brancht}{t_b}
\newcommand{\plant}{N}
\newcommand{\tte}{TTL}

\newcommand{\param}{\theta}
\newcommand{\paramest}{\hat{\theta}}
\newcommand{\pdyn}{f_L}
\newcommand{\belief}{b}

\newcommand{\jointstate}{z}
\newcommand{\jointdyn}{f}

\newcommand{\rdyn}{f_R}
\newcommand{\rstate}{x_R}

\newcommand{\rctrl}{u_R}
\newcommand{\rxdim}{n_R}

\usepackage{multicol}
\usepackage[bookmarks=true]{hyperref}

\usepackage{amsfonts}
\usepackage{algorithmic}
\usepackage[ruled,vlined,titlenotnumbered]{algorithm2e} 
\usepackage{amsmath}
\usepackage{dsfont}
\usepackage{subcaption}
\usepackage{multicol}
\usepackage[pdftex]{graphicx}   
\usepackage{graphicx}
\usepackage{bbm}
\usepackage{mathtools}
\usepackage{dsfont}

\usepackage{booktabs}
\usepackage{multirow}
\usepackage[table,xcdraw]{xcolor}

\graphicspath{{./figs/}}    
\DeclareGraphicsExtensions{.pdf,.png}  

\usepackage[textfont=md,font=footnotesize]{caption}

\usepackage[left=55pt, right=55pt, top=55pt, bottom=58pt]{geometry} 

\IEEEoverridecommandlockouts
\usepackage{balance}

\usepackage[%
    style=numeric-comp,
    sorting=none,
    backend=biber,
    sortcites=true,
    doi=false,
    firstinits=true,
    hyperref,
    isbn=false,
    eprint=false,
    maxcitenames=3, 
    minbibnames=3, 
    maxbibnames=4, 
    block=none]
    {biblatex}
    
    \renewbibmacro{in:}{}
    \AtEveryBibitem{%
  	\clearlist{language}%
  	\clearfield{pages}%
	}
\bibliography{references}
\usepackage{lipsum}

\definecolor{planning_color}{RGB}{69, 174, 254}    
\definecolor{prediction_color}{RGB}{255, 116, 190}
    
\newcommand{\example}[1]%
{
\textbf{Running example:}
\textit{#1}
}

\usepackage[normalem]{ulem}
\usepackage{ifthen}
\newboolean{include-notes}
\setboolean{include-notes}{false}

\newcommand{\adedit}[1]{\ifthenelse{\boolean{include-notes}}{\textcolor{magenta}{{AD (edit):#1}}}{}}

\begin{document}

\title{Analyzing Human Models that Adapt Online}

\author{Andrea Bajcsy$^{1}$, Anand Siththaranjan, Claire J. Tomlin, Anca D. Dragan
\thanks{$^{1}$Authors are with EECS at University of California, Berkeley: \{abajcsy, anandsranjan, tomlin, anca\}@berkeley.edu}
}

\maketitle
\thispagestyle{empty}
\pagestyle{empty}


\begin{abstract}
Predictive human models often need to adapt their parameters \textit{online} from human data. This raises previously ignored safety-related questions for robots relying on these models such as \textit{what} the model could learn online and \textit{how quickly} could it learn it. For instance, when will the robot have a confident estimate in a nearby human's goal? Or, what parameter initializations guarantee that the robot can learn the human's preferences in a finite number of observations? To answer such analysis questions, our key idea is to model the robot's learning algorithm as a dynamical system where the state is the current model parameter estimate and the control is the human data the robot observes. This enables us to leverage tools from reachability analysis and optimal control to compute the set of hypotheses the robot could learn in finite time, as well as the worst and best-case time it takes to learn them. We demonstrate the utility of our analysis tool in four human-robot domains, including autonomous driving and indoor navigation.
\end{abstract}


\section{Introduction}


Robots rely on predictive models of human behavior in order to plan safe and efficient motions around people. Because people vary in their intentions, preferences, and styles of behavior, these models must adapt online upon observing a specific person. For instance, a robot may use Bayesian inference to update its belief about the location a pedestrian is walking to \cite{ziebart2009planning, jain2018recursive, kitani2012activity}, or use online gradient descent to update parameters corresponding to a human's proxemics preferences \cite{bajcsy2017learning, bobu2018learning} or parameters of a neural network which predicts how a human reaches for objects \cite{cheng2019human}. 

Enabling robots to adapt their human models online is necessary and beneficial, but it also raises safety challenges. The human model can now change significantly in light of new data--the human state/actions--that the robot observes. 
What the model parameters change to and how quickly they change depends both on the robot's learning algorithm and how the human actually behaves. Some parameter initializations will be conducive to better learning. Some human behaviors may be ambiguous under the model and result in slower learning. These all have significant implications for the safety and efficiency of the robot's decision-making.

Adaptive models thus raise several questions. First, what is the \emph{worst-case time} it can take the robot to learn the correct model parameters (whatever they may be)? With this, the robot can, for instance, make contingency plans that safeguard against all intents until this worst-case time but then branch on the true parameter value afterwards. In contrast, what is the \emph{best-case time} to learn? Here, a robot can, for example, gain insights about which locations in a room make learning from nearby humans so challenging that even in the best-case it still takes too long to estimate the human's intent. Relatedly, we can also ask what \emph{observations}  lead to the \emph{fastest} or \emph{slowest learning}, thereby producing legible or deceptive motions. Finally, what \textit{initialization} should the robot use, so that we can guarantee that the robot learns the correct parameters-- regardless of what they might be--in a finite number of observations? These questions amount to knowing \textit{what} the robot can learn and in \textit{how much time}, and thus far have been implicitly raised but received little attention.

In this work, our key idea is that we can answer these analysis questions by posing them as \emph{reachability} queries. 
To achieve this, 
\emph{we model the robot's learning algorithm as a \emph{dynamical system}, where the state is the current parameter \emph{estimate}, and the control is the human \emph{data} the robot observes. }
Answering what model parameters can the robot learn and in how much time reduces to answering \textit{when} and \textit{what} states our dynamical system can reach. 


In our formulation, a designer can instantiate a parameterized model of the human (e.g. human driving a car may drive straight or take a left turn at an intersection), an online learning algorithm initialization (e.g. a uniform prior over the human taking a left or driving straight in a Bayesian learning setup), and a desired level of confidence in a particular hypothesis (e.g. acquire high probability on the human taking a left turn), and finally a measurement frequency with which the robot receives data about the human (e.g. human observations are acquired at 10 Hz). 

With the components from above, answering analysis questions amounts to solving an optimal control problem whose solution determines which human observations ``steer'' the learning system into desired states. To determine what our robot could possibly learn, we start our dynamical system with the current estimate of the parameter and solve \textit{forward in time} for the set of hypotheses that are reachable in finite time and observations. We can also compute the minimum (or maximum) time to learn a parameter by solving an optimal control problem \textit{backwards in time} for the earliest time at which there exists a sequence of data points that steer the learning system to the desired parameter. 

We demonstrate a variety of use cases for our analysis tool, answering these questions for an autonomous car operating at an intersection near a human-driven vehicle with unknown intention, and a navigation task in a bookstore environment around a human with unknown goals or proxemics preferences. While in this paper we focus on low-dimensional parametrizations (e.g. unknown human goal), we are encouraged by this first-of-its kind tool for analyzing adaptive human models and are excited to explore approximate DP applications that can extend our tool to higher dimensions.

\section{Related Work}


\smallskip
\noindent\textbf{Learning from human data.} Learning from human data has become increasingly popular in robotics, enabling robots to effectively predict human motion like walking or reaching \cite{ziebart2009planning, kitani2012activity, javdani2015shared, ravichandar2015human}, to learn human preferences from physical interaction \cite{bajcsy2017learning, bobu2018learning}, or to estimate the fidelity of a predictive model \cite{fisac2018probabilistically, fridovich2020confidence}. While research on analyzing algorithms and verifying models \emph{already} learned \textit{offline} from human data has garnered some interest \cite{brown2017toward, sadigh2014data}, to date there has not been research analyzing what models \emph{could} learn. 

\smallskip
\noindent\textbf{Analysis of learning algorithms.}
The advent of modern machine learning has spurred a resurgence of interest in analyzing learning algorithms. While there is a long history of relating ODEs to optimization methods \cite{moore1996optimization}, recently, control theoretic tools such as control Lyapunov functions and reachability analysis have been used to prove convergence and safety properties of gradient-based learning methods \cite{wilson2016lyapunov, laborde2020lyapunov, shi2018understanding}, POMDPs \cite{ahmadi2019control}, and neural networks \cite{ruan2018reachability}. While related, our work focuses not only on analyzing learning algorithms which use human data but also leverages control tools which have not yet been studied in learning contexts and have the ability to compute time-to-reach properties. 


\smallskip
\noindent\textbf{Verification \& control of dynamical systems.}
The control theory and formal methods communities have a rich history of verifying the behavior of dynamical systems. In contrast to the above learning algorithm analyses, these approaches not only focus on convergence and safety, but also on properties such as minimum time-to-reach \cite{yang2013one}, robustness to disturbances \cite{bansal2017hamilton}, and controller synthesis \cite{lygeros2004reachability}. This research lays the foundation for our work wherein we model learning from human data as a dynamical system and we answer analysis queries via solving optimal control problems. 
 

\section{Problem Formulation \& Solution}


\subsection{States \& Dynamics}
The robot observes data in the form of state-action pairs $(\hstate, \hctrl)$. Let the human's discrete-time dynamics be:
    \begin{equation}
        \hstate^{t+\dt} = \hdyn(\hstate^t, \hctrl^t)  
    \end{equation} 
where $\hstate \in \mathcal{X}$ and $\hctrl \in \mathcal{U}$. Additionally, $\dt$ represents the frequency with which our robot observes data about the human (e.g. frequency of the state estimator).

Let $\param \in \Theta$ denote the unknown human model parameter where $\Theta$ is a discrete set of values that $\param$ can take on. This parameter could denote a variety of unknown aspects governing human behavior, from how passive or aggressive a person drives \cite{sadigh2016information}, what locations in a room they are moving towards \cite{ziebart2009planning}, how optimally the person behaves \cite{fridovich2020confidence}, or even the kind of visual cues they pay attention to in a scene \cite{kitani2012activity}. 

The robot uses an online learning algorithm to estimate the human's intent given a sequence of observations over time. Our key idea in this work is to view the robot's learning algorithm as a dynamical system where the parameter estimate is treated as state and the human's data is control input. Let $\paramest$ be the learning algorithm's estimate. This could be, for example, a point estimate of $\param$ or the belief $\belief(\param)$.

Let the discrete-time dynamics of a learning algorithm be
    \begin{equation}
        \paramest^{t+\dt} = \pdyn(\paramest^t, \hstate^t, \hctrl^t)
        \label{eq:dyn_learning}
    \end{equation} 
where $\pdyn: \mathcal{E} \times \mathcal{X} \times \mathcal{U} \rightarrow \mathcal{E}$ is a single update of $\paramest$ given ($\hstate^t, \hctrl^t$) and $\paramest \in \mathcal{E}$ is the space of estimates (e.g. $\mathcal{E} = \Theta$ for a point-estimator or $\mathcal{E} = [0,1]^{|\Theta|-1}$ for the full belief).

To capture the joint evolution between the learning algorithm's estimate of the human's intent parameter and the human's possible behavior, we consider the joint state: 
$
\jointstate := [\hstate,~\paramest]^\top \in \mathcal{Z} = \mathcal{X} \times \mathcal{E}.
$
The human data -- the actions the human takes -- evolve both the physical human dynamics and the learning dynamics. Thus, we would like to analyze the joint dynamical system

\small
    \begin{equation}
        \jointstate^{t + \dt} = \jointdyn(\jointstate^t, \hctrl^t)  := 
        \begin{bmatrix}
        \hdyn(\hstate^t, \hctrl^t) \\
        \pdyn(\paramest^t, \hstate^t, \hctrl^t)
        \end{bmatrix}.
    \end{equation}
\normalsize
Note that this should not be confused with the robot being able to control the human's actions. This joint system representation allows us to answer reachablility questions about which estimates are ``learnable'' under different restrictions on the type of data the human could produce.


\subsection{Solution Method}


Equipped with our joint dynamics which describe the evolution of the human's state and the learning algorithm, we can now formulate an optimal control problem whose solution captures which human observations ``steer'' the learning system into desired states. We build on our framework from \cite{bansal2020hjpred, bajcsy2020robust} and adapt it for analyzing general discrete-time algorithms that learn online from human data.

We define the discrete-time optimal control problem and its associated value function as
    \begin{equation}
        V(\jointstate) \coloneqq \min_{\mathbf{\hctrl}} \min_{t \in \{0, \dt, \hdots, T\}} l(\xi^t(\jointstate, \mathbf{\hctrl}))
        \label{eq:value_fun_def}
    \end{equation}
where $\mathbf{\hctrl}=[\hctrl^0, \hdots,\hctrl^{T-1}]^\top$ is a sequence of controls over the time horizon and $\xi^t(\jointstate, \mathbf{\hctrl})$ is the joint state achieved by applying $\mathbf{\hctrl}$. Here, the function $l(\cdot)$ encodes the distance between the current system state and the desired values of the parameters we seek to estimate.
Different analysis questions simply become different instantiations of this function, as we describe in Sec.~\ref{sec:encoding_analysis_questions}. Intuitively, this value function captures the \textit{closest} our system ever gets to the target states as measured by the signed distance function $l$.  

This minimum-payoff optimal control problem can be solved via the principle of dynamic programming \cite{mitchell2005time}. Specifically, for a finite horizon $t \in \{0, \dt, \hdots, T\}$, the discrete-time time-dependent terminal-value Hamilton-Jacobi-Bellman variational inequality \cite{margellos2011hamilton, fisac2015reach} is:
    \begin{align}
        V^t(\jointstate) &= \min \Big\{l(\jointstate), \min_{\hctrl \in \mathcal{U}^t} V^{t + \dt}(\jointdyn(\jointstate, \hctrl)) \Big\}, \nonumber \\
        V^T(\jointstate) &= l(\jointstate), ~~\forall \jointstate \in \mathcal{Z}
        \label{eq:discrete_time_hjb}
    \end{align}
where $\mathcal{U}^t$ is the set of allowable actions--and therefore data--the human can generate (described in detail in Sec.~\ref{sec:encoding_analysis_questions}). Intuitively, this value function definition can be thought of as analogous to the discrete-time Bellman equation with discount factor $\gamma = 1$, no running cost, and only terminal cost $l(\cdot)$. 


Given this time-dependent value function and a candidate initial condition of the joint state $\jointstate^0$, we can extract two important quantities. First, the optimal control at $\jointstate$ is:
\begin{equation}
    \hctrl^t(\jointstate) = \arg \min_{\hctrl \in \mathcal{U}^t} V^{t + \dt}(\jointdyn(\jointstate, \hctrl)). 
    \label{eq:uopt_compute}
\end{equation}
Secondly, we define the \textit{time-to-learn} (\tte) as the earliest time when our system will reaches the target parameters we seek to learn starting from $\jointstate^0$. More formally, this \tte:
\begin{equation}
    \tte = \min \{t: V^{T-t}(\jointstate^0) \leq 0\}.
    \label{eq:tte_compute}
\end{equation}

\section{Encoding Analysis Questions} 
\label{sec:encoding_analysis_questions}



We can now turn our attention to mathematically encoding the analysis questions we are interested in answering. The key components that enable us to encode analysis questions in Eq.~\eqref{eq:discrete_time_hjb} are (1) the target set of states $\mathcal{L}$ which is encoded via $l(\cdot)$, (2) the set of human actions, $\mathcal{U}^t$, and (3) the strategy of the human (if they are minimizing or maximizing value). 

\smallskip
\noindent\textbf{Time-to-learn (\tte) queries.} 
Computing the best and worse-case time to learn (\tte) depends not only on the learning algorithm, but also on the value of the parameter we are trying to estimate and the type of data the robot could observe. For simplicity, we first consider computing the $\tte$~for a specific parameter value, $\param^*$, and later discuss how to integrate multiple $\tte$~estimates. 


We can mathematically embed the objective of learning $\param^*$ in a target set defined in joint state space: for example, 
$\mathcal{L} = \{\jointstate : \hstate \in \mathcal{X}, ||\paramest - \param^*|| \leq \epsilon \}$. Intuitively, these target sets encode that we want our parameter estimate to be close to the true value, but we do not care where the human ends up in physical state-space as long as we have estimated the parameter well. Defining $\mathcal{L}$ to be a closed set in $\mathcal{Z}$ allows us define a function $l(\jointstate): \mathcal{Z} \rightarrow \mathbb{R}$ such that $\mathcal{L} = \{\jointstate : l(\jointstate) \leq 0\}$. For example, the signed distance function to $\mathcal{L}$ satisfies this property. This function $l(z)$ serves as the cost function for our optimal control problem from \eqref{eq:value_fun_def}.
 
Next, how quickly the robot learns also depends on the possible data about the human the robot observes. Since the observed data is the human behavior, we must define the set of controls $\mathcal{U}^t$ that the human could possibly generate at each time step. In general this set can be chosen in a variety of ways. One of the most straightforward sets is to assume that the robot could observe the human taking \textit{any} action, i.e. $\mathcal{U}^t = \mathcal{U}$. Allowing the human to take any action--and therefore produce any data--leads to very a conservative worst-case \tte, since it allows for the human to abandon any true intent and act purely adversarially. While type of analysis may be desirable in some scenarios, a potentially more realistic set of observations could be $\mathcal{U}^t = \{\hctrl : P(\hctrl \mid \hstate; \param^*) \geq \delta\}$ where $P(\hctrl \mid \hstate; \param^*)$ encodes a state and intent-conditioned action distribution. By varying $\delta \in [0,1]$, the human is allowed to generate more or less sub-optimal data with respect to the intent-driven model. 

Finally, the human's strategy in the control problem determines if we obtain best or worst-case learning estimates. 

\smallskip

\noindent\textbf{Best-case \tte~($\min$).} Since our target set $\mathcal{L}$ encodes the true human intent parameter that the robot wishes to learn, modelling the human as \textit{minimizing} the value which is propagated from $l$ in Eq.~\ref{eq:discrete_time_hjb} encodes a cooperative human. That is, the human is generating data in an attempt to \textit{help} the robot learn their true intent parameter. This enables us to extract the best-case $\tte_{\param^*}$ via Eq.~\eqref{eq:tte_compute}. When interested in the best-case time to learn \textit{any} $\param^* \in \Theta$, we can obtain a conservative best-case $\tte$~via $\max_{\param^* \in \Theta} \tte_{\param^*}$.

\smallskip

\noindent\textbf{Worst-case \tte~($\max$).} Alternatively, by modelling the human as \textit{maximizing} the value, we can easily encode adversarial behavior where the human is trying to prevent the robot from learning their true $\param$ for as long as possible. This enables us to extract the worst-case $\tte_{\param^*}$ via Eq.~\eqref{eq:tte_compute}. Similarly to above, we obtain an upper-bound on learning any $\param^* \in \Theta$ by computing $\max_{\param^* \in \Theta} \tte_{\param^*}$.

\subsection{Reachable parameter queries.}
 Computing the set of forward reachable parameters given an initial estimate follows a similar setup as the $\tte$ queries. The main difference comes from $\mathcal{L}$, which now encodes the initial joint state. For example, $\mathcal{L} = \{\jointstate : \hstate = \hstate^0, \paramest = \paramest^0\}$. Now, the sub-zero level set of $V$ in Eq.~\eqref{eq:discrete_time_hjb}, encodes the set of states our system can \textit{reach} in $T$ time. 

\section{Use Cases of Our Analysis Tool} \label{sec:results}
We now demonstrate a variety of use cases for our tool, ranging from autonomous driving to gradient-based learning. 


\subsection{Synthesizing Safe \& Efficient Contingency Planners}
\label{subsec:contingency_plan}

\begin{figure} [t!]
    \centering
    \includegraphics[width=\columnwidth]{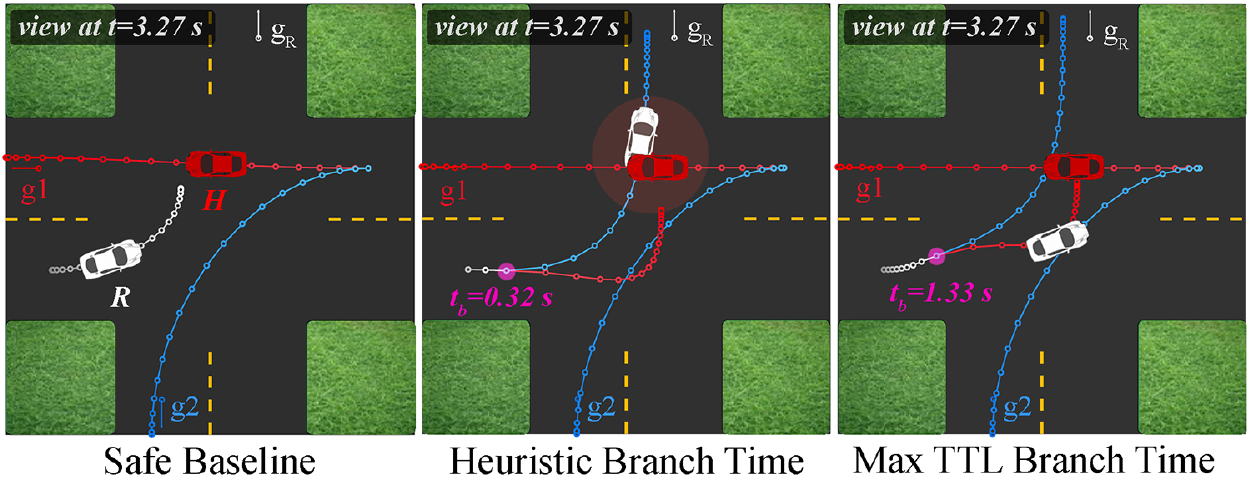}
    \caption{(left) Robot safeguards against both hypotheses, straight and left, for entire planning horizon, (center) A heuristically chosen branch time for the contingency planner doesn't allow the robot to observe enough human data, leading it to collision, (right) Contingency planner branches at the max~\tte~computed via our method, enabling a safe but efficient plan.}
    \label{fig:driving_contingency}
    \vspace{-1em}
\end{figure}

Motion planners for autonomous vehicles often face uncertainty in how human-driven vehicles will behave. Consider the scenario where an autonomous vehicle is attempting to turn left at an unsignalized four-way intersection and there is a human-driven vehicle in the opposing lane (see Fig.~\ref{fig:driving_contingency}). The autonomous vehicle has uncertainty about whether the human will turn left or go straight (goal 1 and goal 2, respectively)-- the outcome of which significantly impacts the autonomous car's ultimate maneuver. This is therefore a prediction problem in which the model's parameter represents the human's maneuver goal: $g := \theta$ and $g \in \Theta = \{g_1, g_2\}$. 

A safe solution to this problem computes a plan for the autonomous vehicle which safeguards against both events during the entire planning horizon, i.e. avoids collisions with the straight and left-turn trajectory. While safe, this approach often yields conservative and inefficient motions (left Fig.~\ref{fig:driving_contingency}).

Alternatively, recent methods have proposed \textit{contingency planning} \cite{hardy2013contingency}. In contingency planning, the robot generates a plan which safeguards against all events for a short horizon which is less than the entire planning horizon: $\brancht < \plant$. After $\brancht$, the motion planner generates $|\Theta|$ contingency plans, each of which safeguards against only a single event. 
The premise of contingency planning is that the robot will know by $\brancht$ which branch to choose, and will no longer need to safeguard against all events. The contingency plan is thus only safe if the robot gains enough certainty to choose which plan to use by the the branching time. If the branching time is too short, then when it comes time for the robot to choose a contingency plan, the robot will still have uncertainty over the human's goal. The robot could now be in a position where no plan  exists which safeguards against \emph{all} events that are still likely (center, Fig.~\ref{fig:driving_contingency}). 

In \cite{hardy2013contingency}, the contingency planning problem is posed as a nonlinear constrained optimization problem which jointly optimizes over the shared segment and the contingency plans while weighting each contingency plan proportional to the belief in that event occurring 
\begin{equation}
    \begin{aligned}
    \arg \min_{\mathbf{\rstate}, \mathbf{\rctrl}} \quad & J_{share}(\rstate^{0:\brancht}) + \sum_{g \in \Theta} b(g)J_{cont}(\rstate^{\brancht+1:\plant}, g)\\
    \textrm{s.t.} \quad & \rstate^{t+1} = \rdyn(\rstate^{t}, \rctrl^t), ~\forall t \in [0, \plant]
    \end{aligned}
\end{equation}
where $\mathbf{\rstate} \in \mathbb{R}^{\rxdim \times \plant}$ denotes a vector containing the robot's planned state trajectory. Here, the robot's dynamics $\rdyn$ are modelled by a 3D Dubins' car where the linear and angular velocity are control inputs and the state is the position and heading. The cost functions $J_{share}$ and $J_{cont}$ encode costs for colliding with static obstacles, reaching the robot's goal, and large changes in acceleration. Additionally, $J_{share}$ penalizes collisions with \textit{all} of the possible outcomes, while $J_{cont}$ penalizes collisions with the human's predicted trajectory towards only the relevant goal, $g_i$. 
A critical design parameter when it comes to the safety of such a motion planning scheme is the choice of branching time, $\brancht$. 

Unfortunately, knowing the future time at which the robot will have certainty about the human's intent is in general challenging. This is where we leverage our analysis framework: we can use it to compute the \emph{worst-case} time it will take the robot to gain certainty in the human intent. 

\begin{table}[t!]
\centering
\resizebox{\columnwidth}{!}{%
\begin{tabular}{@{}l
>{\columncolor[HTML]{EFEFEF}}l ll@{}}
\toprule
& \cellcolor[HTML]{EFEFEF}Prior & \cellcolor[HTML]{C0C0C0}Efficiency (dist to robot's goal) & \cellcolor[HTML]{C0C0C0}Safety (min dist between cars) \\ \midrule
\multicolumn{1}{l|}{\cellcolor[HTML]{C0C0C0}Safeguard Both}                                     & n/a                           &     6.88 m (1.14)       &   0.73 m (0.95)   \\ \midrule
\multicolumn{1}{l|}{\cellcolor[HTML]{C0C0C0}}                                                   & Correct                       &     5.13 m (1.89)       &   0.19 m (0.21)   \\
\multicolumn{1}{l|}{\cellcolor[HTML]{C0C0C0}}                                                   & Incorrect                     &     5.13 m (1.89)       &   -1.55 m (0.99)     \\
\multicolumn{1}{l|}{\multirow{-3}{*}{\cellcolor[HTML]{C0C0C0}Heuristic (0.3265 s)}}             & Uniform                       &     6.10 m (0.79)       &   -0.76 m (1.19)      \\ \midrule
\multicolumn{1}{l|}{\cellcolor[HTML]{C0C0C0}}                                                   & Correct                       &     2.33 m (2.29)       &   0.58 m (0.77)       \\
\multicolumn{1}{l|}{\cellcolor[HTML]{C0C0C0}}                                                   & Incorrect                     &     5.25 m (2.88)       &   0.54 m (0.88)        \\
\multicolumn{1}{l|}{\multirow{-3}{*}{\cellcolor[HTML]{C0C0C0}Max TTL (ours)}}                   & Uniform                       &     3.42 m (3.13)       &   0.55 m (0.77)        \\ \bottomrule
\end{tabular}
}
\caption{Autonomous driving experiment results shown averaged across initial conditions and ground-truth human goals. Mean efficiency and safety metrics are reported in each row and standard deviation in parenthesis.}
\label{tab:driving_results}
\vspace{-2em}
\end{table}

\noindent\textbf{Human Dynamics and Intent Model.} Let the human-driven vehicle be modelled as a 3D Dubins' car with planar position and heading as state and discrete-time dynamics as
$
    \hstate^{t+\dt} = \hstate^t + \dt
    [v_H\cos(\phi), ~
    v_H\sin(\phi), ~
    \hctrl]^\top
$
where the human's control is angular velocity, $\hctrl \in \{-3.5, 0, 3.5 \}~rad/s$ and is driving with a fixed speed $v = 6~m/s$.
The human is modelled as choosing actions via the noisily-rational model \cite{baker2007goal}: $P(\hctrl \mid \hstate, g) \propto e^{Q(\hstate, \hctrl; g)}$ where $Q(\hstate, \hctrl;g)$ encodes the state-action value for the human's driving goal.

\smallskip

\noindent\textbf{Robot Learning Algorithm.} The robot learns the human intent by maintaining and updating a Bayesian belief over $g$. Since $g$ is discrete, the robot can only maintain $|\Theta| - 1 = 1$ probabilities. Without loss of generality, let the robot update $\belief(g = g_1) \coloneqq \paramest$. The learning dynamics $\pdyn$ in Eq.~\eqref{eq:dyn_learning} are
\begin{equation}
    \pdyn(\belief^t(g_1), \hstate^t, \hctrl^t) := \frac{1}{Z}P(\hctrl^t \mid \hstate^t, g_1)\belief^t(g_1)
    \label{eq:driving_belief_update}
\end{equation}
where $Z$ is the normalizer. 

\smallskip

\noindent \textbf{Joint State \& Dynamics.} The joint state is $\jointstate = [\hstate, \belief(g_1)]^\top$ and the joint dynamics are the stacked dynamics equations from above.  
We use $\dt = 0.0891 s$ in all simulations. 

\smallskip

\noindent\textbf{Target Set.} To determine the maximum time it will take our robot to estimate that the human is going forward ($g_1$) or left ($g_2$), we define two target sets in our joint state space:
\begin{align}
    \mathcal{L}_{g_1} &= \{ \jointstate : \hstate \in \mathcal{X}, b(g_1) \geq 0.9\}
    \label{eq:g1_target}\\ 
    \mathcal{L}_{g_2} &= \{ \jointstate : \hstate \in \mathcal{X}, 1-b(g_1) \geq 0.9\}
    \label{eq:g2_target}
\end{align}
Each of these sets encodes that the robot must be at least 90\% confident in the human driving forwards or turning left. 

\smallskip

\noindent\textbf{Computing the Worst-Case Time-to-Learn (\tte).}  Here, we model the human as adversarial and therefore use a $\max$ in the inner value function update from Eq.~\eqref{eq:discrete_time_hjb}. To ensure that while the person is being adversarial, they have to eventually complete their maneuver, we restrict the set of controls to $\mathcal{U}^t = \{\hctrl : P(\hctrl \mid \hstate, g^*) \geq 0.27\}$ where $g^*$ is equal to the human intent which is being analyzed. 
Over a horizon of $T = 1.7820 s$, we perform two value function computations via Eq.~\eqref{eq:discrete_time_hjb} and compute worst-case \tte~ for $g_1$ and $g_2$ by searching backwards in time for the earliest time at which the human's initial state and the robot's initial prior appears in the sub-zero level set of $V^t(\jointstate)$ (as in Eq.~\eqref{eq:tte_compute}) to obtain $\tte_{g_1}$ and $\tte_{g_2}$.  
Finally, to determine the final safe branching time, we want to safeguard against the hypothesis which takes the longest to estimate confidently. Thus, let $\brancht = \max \{ \tte_{g_1}, \tte_{g_2} \}.$

\smallskip 

\noindent\textbf{Results.} We ran a series of simulations comparing the safe motion planner, a heuristically-chosen branching time (comparable to \cite{hardy2013contingency}), and our maximum \tte~branching time contingency planners. For all planners, we varied the initial velocity of the human and robot cars (stopped, moving slowly, or moving quickly) as well as the two possible goals the simulated human was actually moving to. For the contingency planners, we also varied the prior to correctly biased to the human's true goal, incorrectly biased, or uniform. 

Table~\ref{tab:driving_results} summarizes the average efficiency and safety metrics over each of these trials. 
Here, the heuristically-chosen branching time branches too early -- this does not allow the robot to collect enough observations about the human's behavior for the robot to make a confident but safe maneuver, which results in collisions when the robot begins with either a uniform or incorrect prior over the human goals. In contrast, the maximum worst-case \tte~safeguards against both events for a longer time horizon during which the robot collects enough observations to make safe and efficiently goal-driven plans even with a uniform or incorrect prior. Note that the heuristic branching time serves to demonstrate how choosing an uninformed $\brancht$ can lead to safety violations. However, our analysis tool should be thought of as complementary to \cite{hardy2013contingency} wherein we can synthesize an informed $\brancht$.

\subsection{Analyzing Confidence Aware Predictors}
\label{subsec:conf_aware}

Next, we consider predictors that introspect on their model confidence, and use our tool to analyze how long it takes such a predictor to detect that its model cannot explain the observed human behavior. We focus on intent-driven predictors, which model human actions as noisily-optimal with respect to cumulative reward, $P(\hctrl \mid \hstate; \beta) \propto e^{\beta Q(\hstate, \hctrl)}$ \cite{baker2007goal, ziebart2008maximum}. Here, the parameter $\beta$ models how optimally the human behaves: high values of $\beta$ model near-optimal behavior, whereas $\beta=0$ removes the influence of the modelled reward on the human's behavior entirely. Recent work \cite{fisac2018probabilistically, fridovich2020confidence, bajcsy2018scalable} proposed that rather than fixing $\beta$, the robot should estimate it. Upon observing human actions that are poorly explained by the reward function, low values of $\beta$ (signaling low model confidence) will be the most likely, and our predictor will make higher-variance predictions, accounting for its inability to explain the human's behavior.

As the human deviates from the model's assumptions and the predictor makes higher variance predictions, the robot must stay further away from the human to avoid colliding with the now larger set of sufficiently likely states. 
This begs the question: when the human doesn't actually optimize the modeled reward, how long will it take the predictor to adapt its $\beta$ and detect that its model confidence is low? 



\begin{figure}[t!]
    \centering
    \begin{subfigure}[t]{0.35\columnwidth}
        \centering
        \includegraphics[height=2in]{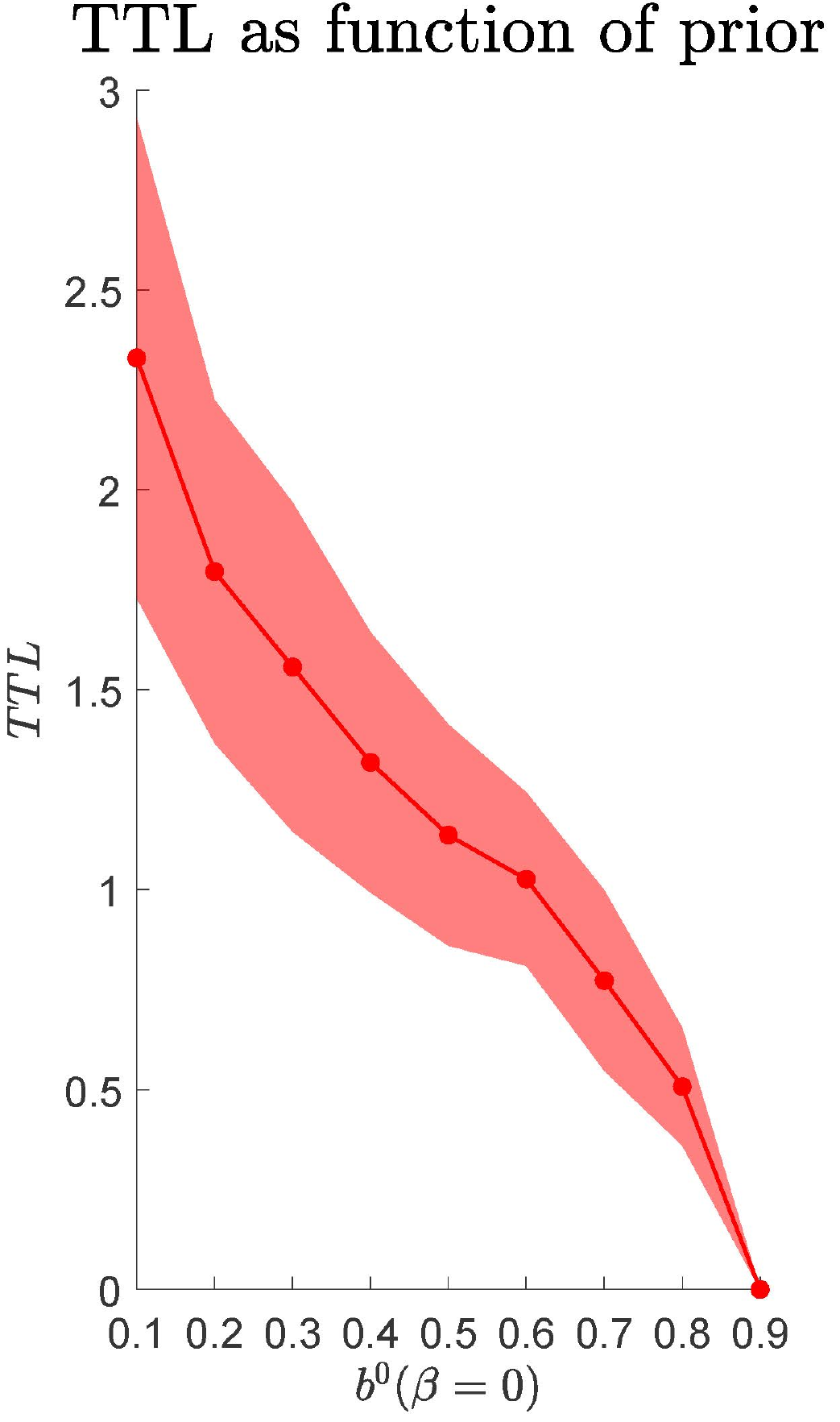}
    \end{subfigure}%
    ~ 
    \begin{subfigure}[t]{0.6\columnwidth}
        \centering
        \includegraphics[height=2in]{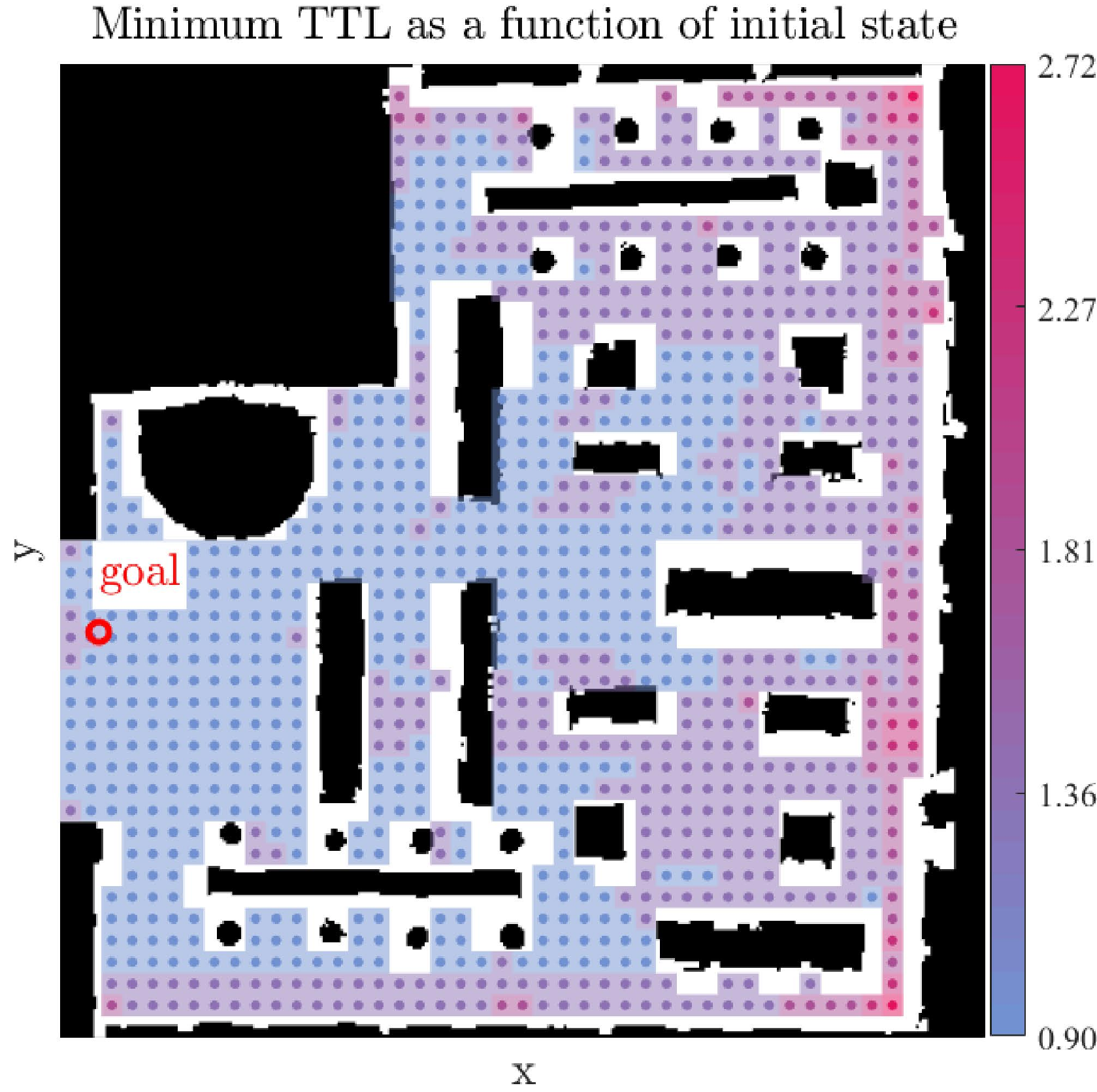}
    \end{subfigure}
    \caption{(left) Minimum \tte~that the human is being unmodelled as a function of the prior. Mean and standard deviation shown in red. (right) Minimum \tte~ as a function of $\hstate$ with a uniform prior. Occupancy map of the bookstore environment (in Fig.~\ref{fig:gradient_initialization}) shown with modelled human goal is red circle and the min \tte~for each initial (x,y) state is shown in a color ranging from blue (\tte=0.9s) to red (\tte=2.72s).}
    \label{fig:conf_results}
    \vspace{-2em}
\end{figure}

Here, 
let the unknown parameter be $\beta \coloneqq \theta$ with $\beta \in \Theta = \{0, 1\}$, signaling low and high model confidence respectively. 

\smallskip 

\noindent\textbf{Human Dynamics and Intent Model.} Let the human be modelled by a simple planar pedestrian model:
$
    \hstate^{t+\dt} = \hstate^t + 
    \dt [v_H\cos(\hctrl),
    v_H\sin(\hctrl),
    ]^\top
$
where the human's control is their heading, $\hctrl \in \{-\pi, \hdots, \pi \}$ and the human is walking at a leisurely speed ($v = 0.6~m/s$) or is stopped ($v = 0$). The human's reward function encourages motion towards the door (shown in red) in the indoor environment (top-down occupancy map shown on right of Fig.~\ref{fig:conf_results}). 

\smallskip

\noindent\textbf{Robot Learning Algorithm.} The robot maintains and updates a Bayesian belief over the confidence parameter $\beta$. Without loss of generality, let the robot explicitly update $\belief(\beta = 0) \coloneqq \paramest$. Thus, $\pdyn$ is identical to \eqref{eq:driving_belief_update} but with $\belief(\beta = 0)$. 

\smallskip

\noindent\textbf{Joint State \& Dynamics.} The joint state is $\jointstate = [\hstate, \belief(\beta = 0)]^\top$ and the joint dynamics are the stacked physical and learning dynamics from above. Finally, we use $\dt = 0.4545~s$. 

\smallskip

\noindent\textbf{Target Set.} We are primarily interested in determining how long it will take our robot to estimate that our model cannot explain the human's behavior ($\beta = 0$). 
Thus, we are interested in answering ``From the prior over the model confidence, what is the \textit{fastest} our robot could learn that the person is behaving in an unmodelled way?'' Our corresponding target set encoding this question is
$
\mathcal{L}_{\beta=0} = \{\jointstate : \hstate \in \mathcal{X}, b(\beta = 0) \geq 0.9\} 
$
where $\epsilon = 0.9$ is our desired confidence. 

\smallskip

\noindent\textbf{Computing the Best-Case Time-to-Learn (\tte). } We seek to compute fastest time to learn we have low confidence in our model, since this is the lower bound on reaction time to unmodelled data. Thus, we use $\min$ over $\mathcal{U}^t$ in the optimization from \eqref{eq:discrete_time_hjb} and we optimize over all data the human could generate, since in the worst case the human is not behaving according to the specified reward function at all. In the following two analyses, the human navigates in a bookstore environment \cite{aws2020environments} whose occupancy map is shown in right of Fig.~\ref{fig:conf_results} and 3D model is shown in Fig.~\ref{fig:gradient_initialization}.

\smallskip

\noindent\textbf{Results: Best-case \tte~as function of \textit{prior}.} We first analyzed the min \tte~a low model confidence as a function of the prior. After computing one backwards reachability computation, we extracted the \tte~for 121 initial $\hstate$ states and 8 levels of the prior. The mean and standard deviation across all initial conditions shown in the left Fig.~\ref{fig:conf_results}. This analysis reveals the added difficulty for the robot to detect its model is wrong if it begins with an optimistic prior.

\noindent\textbf{Results: Best-case \tte~as function of \textit{initial human state}.} Right of Fig.~\ref{fig:conf_results} shows how the best-case \tte~varies as function of the initial $\hstate$ in a complex environment. Here we fix a uniform prior over the model confidence and use the same value function computed from above to query for the $\tte$~ for 1,010 initial human states. Interestingly, this analysis demonstrates that the best-case $\tte$ is largely impacted by the constraints of the physical environment. If the human begins in the open-space near the door, learning that they do not want to move to the door is easy since the human can directly move away from the door to indicate this mismatch. However, if the human begins in a heavily constrained part of the environment such as the lower right-hand corner, all collision-free actions appear ambiguous under the robot's likelihood model; that is, moving left or up could either indicate the human intends to move to the door or they intend to move in a completely different direction.  

A few interesting takeaways for designers of robot motion planners which rely on confidence-aware models include: (1) if the robot does not re-plan faster than this best-case \tte~then the robot will not be able to react quickly enough to the human's unmodelled behavior and (2) the robot should remain more cautious around the human in constrained parts of the environment due to the increased learning uncertainty.


\subsection{Generating Legible \& Deceptive Behaviors} 

\begin{figure} [t!]
    \centering
    \includegraphics[width=1\columnwidth]{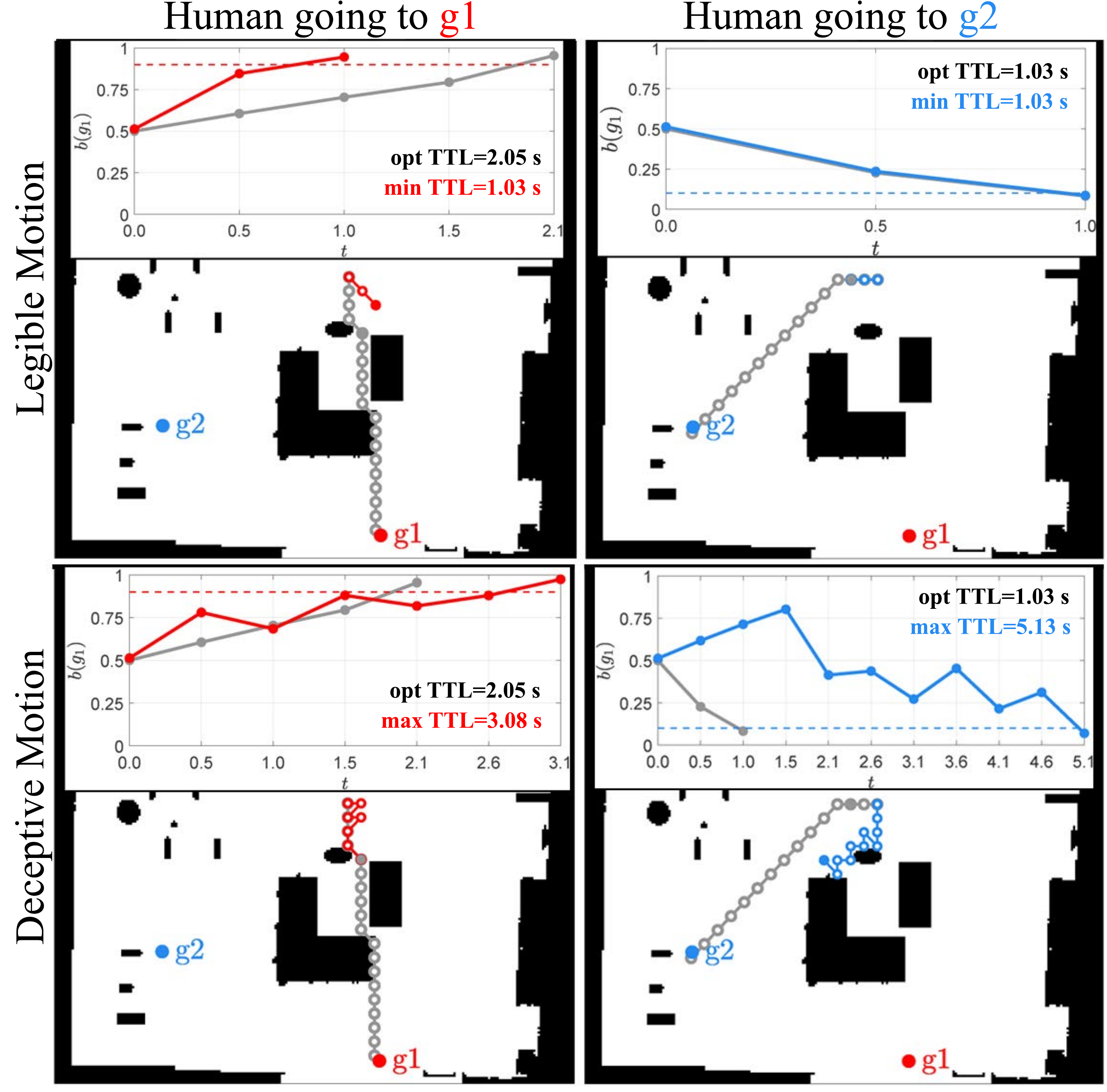} 
    \caption{Legible and deceptive behaviors as synthesized by our analysis tool--shown in bright red or bright blue. The optimal policy for each goal is shown in grey. The estimate of the goal over time for the legible and deceptive behaviors is contrasted with the optimal policy in the inset figures. }
    \label{fig:legible_deceptive}
    \vspace{-1em}
\end{figure}

So far we demonstrated how to compute the worst and best-case learning times. We now showcase how our analysis tool can also synthesize the behavior which led to the fastest or slowest learning by our (robot) observer. 

We use the same planar pedestrian model from \ref{subsec:conf_aware}. In our running example, the human is walking around a living room (occupancy map shown in Fig.~\ref{fig:legible_deceptive}) and the robot has uncertainty about which location the human is navigating to. Let the uncertain human model parameter be $g := \param$ and $g \in \Theta = \{g_1, g_2\}$. The robot uses a noisily-rational model as above, parameterized by $g$, and learns via a Bayesian update. 

We perform four reachability computations, two of which have a high-confidence in $g_1$ target set as in \eqref{eq:g1_target} and two of which have a high-confidence in $g_2$ target set (like \eqref{eq:g2_target}). However, for each pair of reachability computations, we perform one computation where the human is \textit{minimizing} the value (i.e. helping robot learning) and the other where the human is \textit{maximizing} the value (i.e. trying to slow down robot learning). In all examples, the human chooses from $\mathcal{U}^t = \{\hctrl : P(\hctrl \mid \hstate, g^*) > 0.15\}$ where $g^*$ the goal being analyzed. The resulting four value functions are used via Eq.~\eqref{eq:uopt_compute} to extract the optimal sequence of human data which lead to best and worst-case learning times for $g_1,g_2$. 

Fig.~\ref{fig:legible_deceptive} visualizes the optimal controls in the bright colored trajectory corresponding to the human's true goal. We contrast this with the optimal only-goal-driven policy in grey. Inset plots show $\belief(g_1)$ over time and the target confidence level plotted as a dashed line. To increase the probability on $g_1$ as fast as possible, the human moving to $g_1$ signals this by quickly moving to their left (instead of moving forward as in the optimal path). This is in line with prior work on legibility \cite{dragan2013generating}, but a potential advantage of our formulation is that the objective of minimizing $\tte$~is task-oriented. While prior work encouraged agents to be legible along the entire trajectory, here the agent is directly minimizing the time to convey the goal, so that the observer can react as quickly as possible. Interestingly, to be deceptive, the human zig-zags to confuse the observer for the longest. 


\subsection{Online Gradient-based Learning from People} 

\begin{figure*}[t!]
    \centering
    \includegraphics[width=\textwidth]{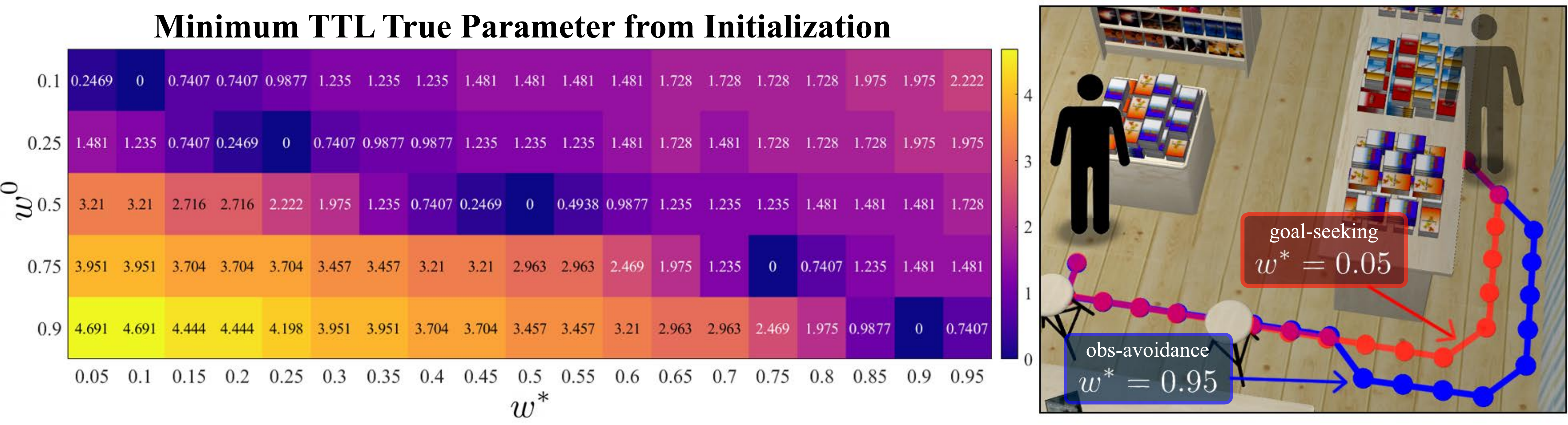}
    \caption{(left) Heatmap of reachable reward weights (x-axis) and their $\tte$~(color values ranging from dark blue: $\tte=0.0s$ to yellow: $\tte=4.7s$) starting from a specific initialization (y-axis). (right) Bookstore environment with faded human-figure denoting the human's initial position: red path is optimal behavior for mainly goal-driven human, blue path is for a human who wants to stay far from obstacles. }
    \vspace{-2em}
    \label{fig:gradient_initialization}
\end{figure*}

Online gradient-based learning algorithms are also useful in many HRI domains \cite{bajcsy2017learning,bobu2018learning,murata2019achieving,yu2018one}. In this case study, we use our analysis tool to determine a parameter initialization which allows the online gradient algorithm to adapt the fastest to any true human intent. As above, the robot is learning a nearby human's reward function by observing their behavior. The human's reward function is modelled as a linear combination of features: $r(\hstate, \hctrl; \param) = \param^\top \mu(\hstate, \hctrl)$. 
Here, $\param = [1-w,~w]^\top$ and $\mu(\hstate, \hctrl) \in \mathbb{R}^2$, encoding the distance between the human and their goal and the distance between the human and obstacles. Adjusting $w$ trades off the human's goal-driven and obstacle-avoidance preferences.

The robot learns about the human's reward via online gradient descent. Thus, the $\pdyn$ from Eq.~\eqref{eq:dyn_learning} are:

\small
\vspace{-0.5em}
\begin{equation}
        \pdyn(\paramest^t, \hstate^t, \hctrl^t) := \paramest^t + \alpha \nabla_{\paramest} F(\hstate^t, \hctrl^t, \paramest^t)
\end{equation}
\normalsize
Maximizing the likelihood of the observed $(\hstate^t, \hctrl^t)$ pair under the maximum entropy distribution \cite{ziebart2008maximum}, we derive a gradient-based update:
$F(\hstate^t, \hctrl^t, \paramest^t) := Q(\hstate^t, \hctrl^t; \paramest^t) - \mathbb{E}_{u \sim P(u \mid \hstate^t; \paramest^t)} \big[ Q(\hstate^t, u; \paramest^t)\big].$
Note: this is the state-action equivalent of learning offline from demonstrations \cite{ziebart2008maximum}.

In this analysis, we solve a \textit{forward reachability} problem where we start our system in the target set $\mathcal{L} = \{\jointstate : \hstate = \hstate^0, \paramest = \paramest^0\}$ where $\hstate^0$ is the current physical state of the human and $\paramest^0$ is a candidate initialization. We compute the set of $\param$'s for which there exists a sequence of observations which evolve the joint system to that $\param$-state in finite time. Here, $\dt = 0.2469~s$ and the total time horizon for which we evolve our system is $T = 7.1605~s$. 
    
Fig.~\ref{fig:gradient_initialization} shows the reachable $\param^* = [1-w^*, w^*]$'s starting from a given $\param^0=[1-w^0, w^0]$. Colors in the heatmap represent the earliest time at which the robot can learn a $\param^*$ starting from a each initialization. Interestingly, if the robot begins with $w^0 = 0.9$ (i.e. human is primarily obstacle-averse) it takes $\sim4.7~s$ to learn that the person is \textit{actually} primarily goal-seeking ($w^* = 0.1$). In contrast, initializing with $w^0 = 0.25$ allows the robot to learn \textit{any} other parameter in $<2.2~s$. Intuitively, this discrepancy in how quickly the robot can learn from an initialization is because of the structure of the environment which in turn affects the gradient update. Since here the person begins in a part of the environment where their direct path is obstructed by obstacles (right Fig.~\ref{fig:gradient_initialization}), they must navigate around the obstacles before they get to the goal. The evidence of the person moving \textit{away} from obstacles makes it difficult to disambiguate if they truly are goal-driven or obstacle-averse. Thus, initializations which bias our estimator towards believing that people are obstacle-averse is a poor choice if we want our robot to learn any other $\theta^*$ quickly.    
    
    



\smallskip
\noindent \textbf{Closing Remarks.} In this work, we leveraged tools from reachability analysis to analyze human models which adapt online. By treating these models as dynamical systems where the estimate is state and the human data is control, we obtain the best and worst-case time to learn, extract the optimal measurements which enable learning, and synthesize good model parameter initializations.



\section*{ACKNOWLEDGMENT}
This research is supported by NSF CAREER, DARPA Assured Autonomy, NSF VeHICal, and SRC CONIX. Andrea Bajcsy is supported by the NSF GRFP. 

\printbibliography

\end{document}